\documentclass{article}

\usepackage{arxiv}

\usepackage{natbib}
\usepackage{graphicx}
\usepackage{amsmath,amssymb,amsfonts}
\usepackage{booktabs}
\usepackage{listings}
\usepackage{xcolor}
\usepackage{float}
\usepackage{adjustbox}
\usepackage{tikz}
\usetikzlibrary{positioning,arrows.meta,calc,fit,backgrounds}
\usepackage{pgfplots}
\pgfplotsset{compat=1.18}

\usepackage{hyperref}
\usepackage[capitalize]{cleveref}

\definecolor{cinput}{RGB}{230,242,255}
\definecolor{cgen}{RGB}{228,245,228}
\definecolor{cckan}{RGB}{242,232,255}
\definecolor{closs}{RGB}{255,235,235}
\definecolor{caux}{RGB}{255,245,225}
\definecolor{cline}{RGB}{55,55,55}

\tikzset{
  >={Latex[length=3mm]},
  flow/.style={-{Latex[length=3mm]}, thick, draw=cline},
  skip/.style={-{Latex[length=3mm]}, thick, dashed, draw=black!70},
  box/.style={draw=cline, rounded corners=2mm, very thick, minimum height=10mm, align=center, font=\small, inner sep=4pt},
  inputbox/.style={box, fill=cinput},
  genbox/.style={box, fill=cgen},
  ckanbox/.style={box, fill=cckan},
  lossbox/.style={box, fill=closs},
  auxbox/.style={box, fill=caux},
  note/.style={font=\footnotesize, align=left}
}



\title{SRGAN-CKAN: Expressive Super-Resolution with Nonlinear Functional Operators under Minimal Resources}

\author{
  Roberto Isai Navarro-Aviña\\
  Cinvestav, Unidad Guadalajara\\
  Av. del Bosque 1145, El Bajio\\
  Zapopan, Jalisco, 45017, Mexico \\
  \texttt{Isai.Navarro@cinvestav.mx} \\
   \And
 Eduardo Said Merin-Martinez \\
  Cinvestav, Unidad Guadalajara\\
  Av. del Bosque 1145, El Bajio\\
  Zapopan, Jalisco, 45017, Mexico \\
  \texttt{eduardo.merin@cinvestav.mx} \\
   \And
 Andres Mendez-Vazquez \\
  Cinvestav, Unidad Guadalajara\\
  Av. del Bosque 1145, El Bajio\\
  Zapopan, Jalisco, 45017, Mexico \\
  \texttt{andres.mendez@cinvestav.mx} \\
  \And
  Eduardo Rodriguez-Tello\\
  Cinvestav, Unidad Tamaulipas\\
  Km. 5.5 Carretera Victoria - Soto La Marina\\
  Victoria, 87139, Tamaulipas, Mexico \\
  \texttt{ertello@cinvestav.mx} \\
}

\begin{document}
\maketitle

%%%%%%%%%%%%%%%%%%%%%%%%%%%%%%%%%%%%%%%
%%%%%%%%%%%%%%%%%%%%%%%%%%%%%%%%%%%%%%%%%%%%%%%%%%%%

\begin{abstract}
    Single-Image Super-Resolution (SISR) aims to reconstruct a High-Resolution (HR) image from a Low-Resolution (LR) observation, a fundamentally ill-posed problem where high-frequency details are severely degraded at large upscaling factors. Recent advances have been driven by transformer-based architectures and diffusion models improve global context modeling and perceptual quality at the cost of increased computational complexity. In contrast, this work focuses on enhancing the expressivity of local operators under minimal resources. We propose SRGAN--CKAN, a hybrid super-resolution framework that integrates Convolutional Kolmogorov--Arnold Networks (CKAN) into an adversarial learning setting reformulating convolution as a nonlinear patch-based transformation. The proposed operator replaces linear local mappings with spline-based functional representations, allowing expressive modeling of complex local structures and high-frequency textures using minimal hardware resources. Experimental results demonstrate that the proposed approach improves perceptual quality while preserving reconstruction fidelity, achieving a favorable balance between distortion-based and perceptual metrics. These results are obtained under constrained computational settings, highlighting the efficiency of the proposed formulation. Overall, this work introduces a complementary direction to existing approaches by improving the representational power of local transformations, providing an efficient and scalable alternative to globally intensive architectures.

\end{abstract}

	%%%%%%%%%%%%%%%%%%%%%%%%%%%%%%%%%%%%%%%%%%%%%%%%%%%%
    \section{Introduction}
    %%%%%%%%%%%%%%%%%%%%%%%%%%%%%%%%%%%%%%%%%%%%%%%%%%%%
    
    Single Image Super-Resolution (SISR) is a classical problem in digital image processing and computer vision that reconstructs a HR image from a single LR observation by recovering high-frequency details lost during image acquisition or transmission. This problem arises in applications such as medical imaging, remote sensing, surveillance, microscopy, and astronomy, where physical or hardware limitations restrict direct acquisition of HR data \citep{dong2014srcnn, kim2016vdsr, srSurveyModern}.
    
    From a signal-processing perspective, SISR is an ill-posed inverse problem \citep{gonzalezwoods, jain1989fundamentals}. The LR image ($I^{LR}$) is typically modeled as a degraded version of the HR image ($I^{HR}$) through blurring, downsampling, and noise:
    $I^{LR} = (I^{HR} * k)\downarrow_s + n,$
    where $*$ denotes spatial convolution with a blur kernel $k$, $\downarrow_s$ is a downsampling operator with scale factor $s$, and $n$ represents additive noise. Because multiple HR images can produce the same LR observation, recovering the original image requires additional priors or learned mappings.
    
    Formally, we denote $I^{HR} \in \mathbb{R}^{3 \times H \times W}$ as the ground-truth high-resolution image and $I^{LR} \in \mathbb{R}^{3 \times h \times w}$ as its low-resolution counterpart, where $H, W$ and $h, w$ represent spatial dimensions with $h = H/s$ and $w = W/s$. The objective is to learn a parametric mapping $G_\theta$ such that: $\hat{I}^{HR} = G_\theta(I^{LR}), \quad \hat{I}^{HR} \approx I^{HR},$ where $\hat{I}^{HR}$ denotes the reconstructed super-resolved image.
    
    Early SISR approaches relied on interpolation methods such as nearest-neighbor, bilinear, and bicubic interpolation \citep{gonzalezwoods}, which are computationally efficient, but produce overly smooth results and fail to recover fine structures. Learning-based methods improved upon these limitations by exploiting self-similarity or external datasets to learn mappings between LR and HR patches \citep{dong2014srcnn, kim2016vdsr}. With the advent of deep learning, Convolutional Neural Networks (CNNs) enabled end-to-end nonlinear mappings, significantly improving reconstruction quality and texture modeling.
    
    A major advance in perceptual super-resolution was introduced by Generative Adversarial Networks (GANs). In particular, SRGAN \citep{ledig2017} demonstrated that combining perceptual loss functions with adversarial training produces visually realistic textures beyond pixel-wise optimization. ESRGAN \citep{esrgan2018} further improved these results through architectural refinements and enhanced loss functions. However, recent studies highlight a fundamental trade-off between distortion-oriented models (optimized for Peak Signal-to-Noise Ratio, PSNR) and perceptual models (optimized for visual realism), as well as challenges in training stability, generalization, and evaluation under adversarial settings \citep{srSurveyModern, goodfellow2014gan, ledig2017}.
    
    From a computational perspective, convolutional operators provide efficient local processing but are inherently limited in their expressive capacity due to fixed linear kernels followed by pointwise nonlinearities. Increasing model expressiveness typically requires deeper or wider architectures, which leads to higher computational and memory costs. This motivates the exploration of alternative operators capable of achieving richer representations without significantly increasing architectural complexity.

\begin{figure*}[ht]
        \centering
        \includegraphics[width=\textwidth]{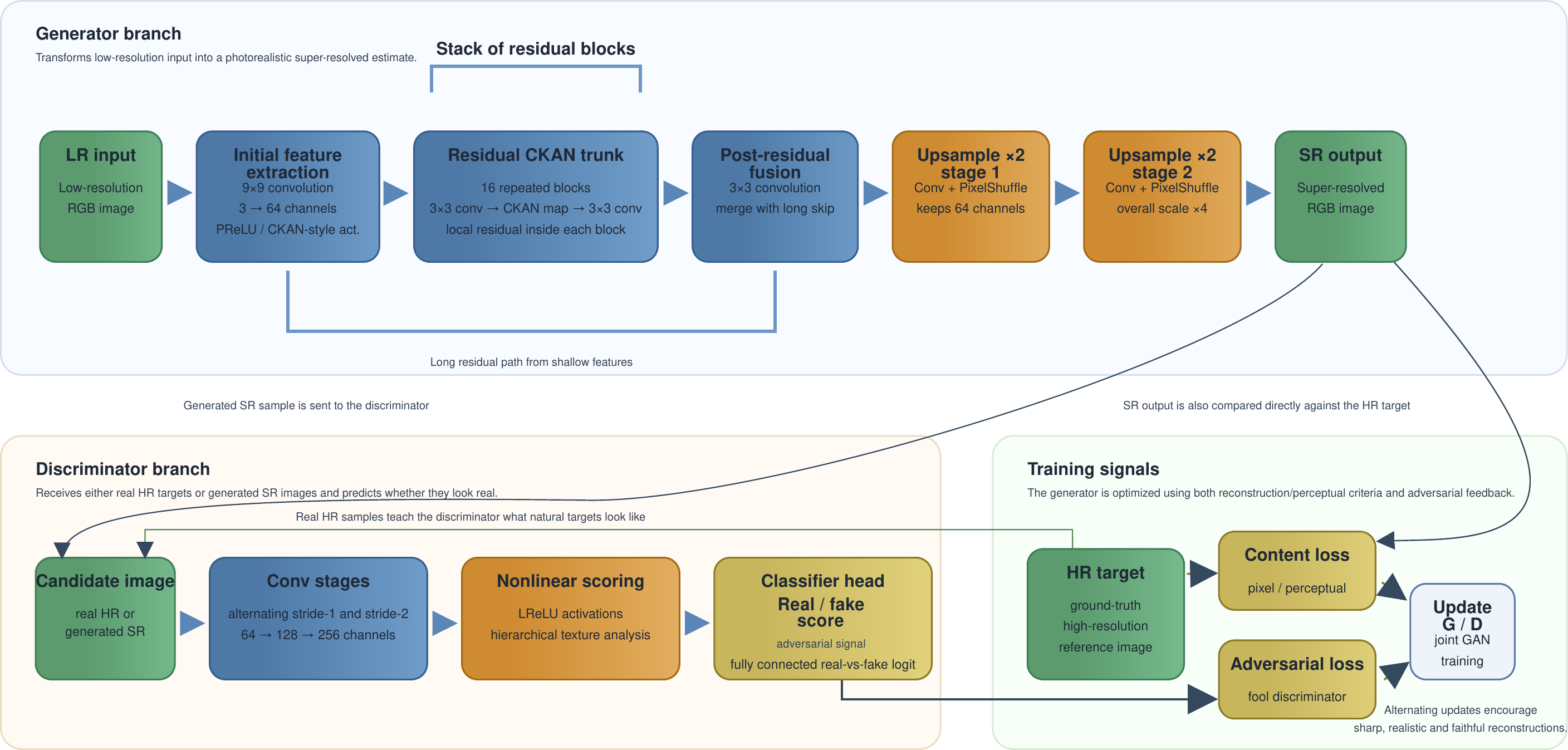}
        \caption{SRGAN-CKAN adversarial interaction (\ref{appendix:srgan_ckan_architecture}). The generator produces super-resolved images using CKAN-based operators, while the discriminator evaluates perceptual realism.}
        \label{fig:srgan_ckan_overview}
    \end{figure*}

    In this work, we propose the SRGAN--CKAN architecture, which integrates the SRGAN framework with operators inspired by Kolmogorov--Arnold Networks (KAN) \citep{kan2024}. The proposed model replaces standard convolutions with a patch-based projection operator that applies learnable nonlinear mappings using spline-based transformations \citep{deboor1993box}, consistent with Kolmogorov--Arnold functional representations \citep{kan2024}. This formulation enables more expressive modeling of local image structures compared to traditional convolution. An overview of the proposed architecture is shown in Fig.~\ref{fig:srgan_ckan_overview}, illustrating the integration of CKAN-based operators within the generator, the discriminator pipeline, and the interaction between adversarial and reconstruction-based training objectives..

    Importantly, the proposed CKAN operator introduces this nonlinear patch-based projection while maintaining a computational complexity comparable to standard convolution under linear projection settings. This allows the model to increase representational capacity without incurring prohibitive computational overhead. A detailed analysis of time and memory complexity is provided in Appendix~\ref{appendix:ComputationaComplexityLBTs-KAN}.
    
    The objective of this work is to use KAN-style operators into super-resolution architectures maintaining a stable adversarial training framework, and the use of minimal resources in hardware\footnote{https://github.com/RINavarro/SRGAN-CKAN}. The proposed approach preserves the SRGAN training paradigm while incorporating nonlinear patch-wise mappings, enabling improved modeling of complex textures under constrained computational resources.

    \begin{figure*}[ht]
        \centering
        \includegraphics[width=\textwidth]{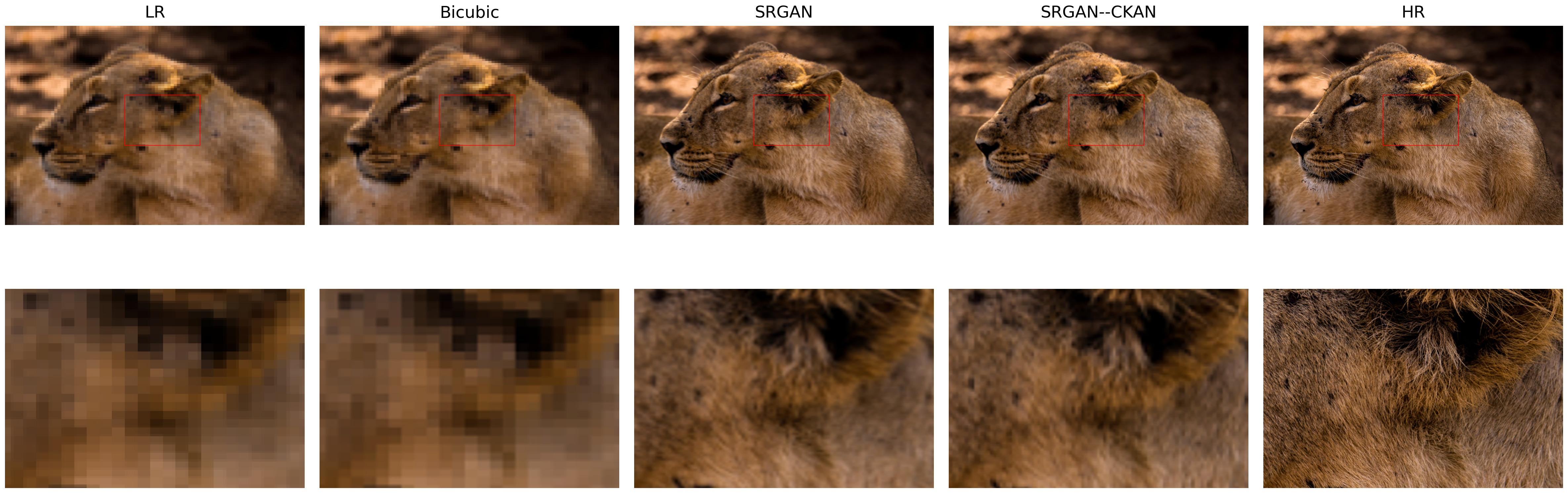}
        \caption{Visual comparison between LR input, bicubic interpolation, conventional SRGAN, the proposed SRGAN--CKAN model, and the HR reference. The zoom-in regions highlight differences in texture reconstruction and artifact suppression.}
        \label{fig:sr_comparison}
    \end{figure*}

    We evaluate the proposed method against representative approaches from the literature, including convolution-based SRGAN variants and perceptual super-resolution models such as SRGAN and ESRGAN, which serve as strong baselines for adversarial image reconstruction. This comparison allows us to assess both reconstruction fidelity and perceptual quality under a controlled experimental setting.
    
    The main contributions of this work are summarized as follows:
    \begin{itemize}
        \item We reformulate convolutional operators within the SRGAN framework using a Kolmogorov--Arnold-inspired projection mechanism (CKAN), enabling a patch-wise functional representation of local image transformations.
        
        \item We design and implement a complete SRGAN--CKAN training pipeline, including supervised pretraining and adversarial fine-tuning under constrained computational resources.
        
        \item We provide a controlled experimental analysis showing that the proposed model preserves reconstruction fidelity while improving perceptual quality, achieving a favorable trade-off between Peak PSNR, Multi-Scale Structural Similarity (MS-SSIM), and Learned Perceptual Image Patch Similarity (LPIPS).
        
        % \item We analyze the stability of adversarial training under the proposed formulation, demonstrating consistent convergence without mode collapse.
    \end{itemize}
    
    The remainder of this paper is organized as follows. Section II reviews related work in super-resolution and adversarial learning. Section III presents the proposed SRGAN--CKAN architecture and its mathematical formulation. Section IV describes the experimental setup and evaluation metrics. Section V discusses the results and comparisons with state-of-the-art methods. Finally, Section VI concludes the paper and outlines future research directions.
    
    %%%%%%%%%%%%%%%%%%%%%%%%%%%%%%%%%%%%%%%%%%%%%%%%%%%%
    \section{Related Work}
    %%%%%%%%%%%%%%%%%%%%%%%%%%%%%%%%%%%%%%%%%%%%%%%%%%%%

   SISR aims to reconstruct a high-resolution image from a degraded low-resolution observation. Due to its ill-posed nature, effective solutions require both strong priors and expressive models capable of capturing complex image structures.

    Early learning-based approaches relied on example-based methods that exploit self-similarity or external datasets to learn mappings between LR and HR patches. With the advent of deep learning, CNNs significantly improved performance by enabling end-to-end nonlinear mappings from $I^{LR}$ to $I^{HR}$. Seminal works such as SRCNN \cite{dong2014srcnn} demonstrated the effectiveness of shallow CNNs, while deeper architectures such as VDSR \cite{kim2016vdsr} and EDSR \cite{lim2017enhanced} further improved reconstruction quality through residual learning and increased model capacity. These methods typically optimize pixel-wise losses, achieving high PSNR but often producing overly smooth results.
    
    To address these limitations, perceptual approaches introduce losses defined in feature space. In particular, Super-Resolution Generative Adversarial Network (SRGAN) \cite{ledig2017} combines adversarial training with perceptual loss to produce visually realistic textures, while ESRGAN \cite{esrgan2018} improves stability and reconstruction quality through architectural and training refinements. These methods aim to generate images closer to the manifold of natural images rather than minimizing pixel-wise error, but introduce challenges related to training stability and the trade-off between perceptual quality and reconstruction fidelity.
    
    More recently, attention mechanisms and transformer-based architectures have been introduced to enhance feature representation. Channel attention models such as RCAN \cite{zhang2018imagesuperresolutionusingdeep} improve performance by adaptively reweighting feature channels, while transformer-based methods such as SwinIR \cite{liang2021swinir} capture long-range dependencies through self-attention mechanisms. These approaches significantly improve representational capacity, but increasing computational complexity.
    
    In parallel, diffusion-based methods have emerged as a powerful paradigm for perceptual super-resolution. Approaches such as SR3 \cite{saharia2021imagesuperresolutioniterativerefinement} reconstruct high-resolution images through iterative refinement processes, achieving remarkable perceptual quality requiring substantial computational resources and inference time.
    
    Despite their success, most super-resolution architectures rely on convolutional operators that compute local responses as linear combinations of features:
    \[
    y_{i,j} = \sum_{c=1}^{C} \sum_{u,v} w_{c,u,v}\, x_{c,\,i+u,\,j+v},
        \]
    where $x \in \mathbb{R}^{C \times H \times W}$ denotes the input feature map with $C$ channels, $w_{c,u,v}$ are the convolutional kernel weights, and $(i,j)$ indexes the spatial location of the output feature map. The indices $(u,v)$ span the local receptive field defined by the kernel size, and $x_{c,\,i+u,\,j+v}$ corresponds to the input value at channel $c$ and spatial position $(i+u, j+v)$. While nonlinear activations are applied across layers, the local transformation itself remains linear, limiting the modeling of highly textured and context-dependent structures.
    
    This observation motivates the exploration of alternative formulations that directly enhance the expressivity of local operators. In contrast to approaches that focus on global context modeling or generative refinement, the CKAN-based operator proposed in this work replaces linear convolutional mappings with nonlinear patch-based functional transformations, enabling more expressive modeling of local image structures while maintaining computational efficiency.
        
	%%%%%%%%%%%%%%%%%%%%%%%%%%%%%%%%%%%%%%%%%%%%%%%%%%%%
    \section{Proposed Method}
    %%%%%%%%%%%%%%%%%%%%%%%%%%%%%%%%%%%%%%%%%%%%%%%%%%%%
    
    Our approach builds upon the SRGAN framework, where a generator $G$ learns a mapping from a low-resolution image $I^{LR}$ to a reconstructed high-resolution image $\hat{I}^{HR}$, while a discriminator $D$ distinguishes between generated images and real high-resolution samples.

    Despite its effectiveness, this framework relies on convolutional operators that are inherently linear at the level of local patches. A convolution can be interpreted as extracting a local region and representing it as a vector $\mathbf{p}_\ell \in \mathbb{R}^{C \cdot k^2}$, leading to $y_{i,j} = \mathbf{w}^\top \mathbf{p}_\ell.$
    
    This formulation reveals that convolution performs a linear projection in patch space. While nonlinearities are introduced across layers, the local transformation itself cannot explicitly model nonlinear interactions within the patch. This limitation becomes critical in super-resolution, where reconstructing high-frequency details requires capturing complex, context-dependent relationships between neighboring pixels.
    
    To address this limitation, we propose the CKAN operator, which replaces linear convolutional kernels with nonlinear mappings defined over local patches. The proposed formulation is grounded in Kolmogorov--Arnold Networks, specifically leveraging the LTBs-KAN framework \cite{merinmartinez2026ltbskanlineartimebsplineskolmogorovarnold}, which enables efficient spline-based functional representations with linear computational complexity, \cref{appendix:ComputationaComplexityLBTs-KAN}.
    
    Under this formulation, each local patch is treated as a vector-valued input, and the transformation is performed through learnable spline-based nonlinear functions, allowing the model to capture richer local interactions while preserving computational efficiency.
    
    \subsection{CKAN Operator}
    
    Given an input feature map $X \in \mathbb{R}^{B \times C \times H \times W}$, where $B$ is the batch size, $C$ denotes the number of input channels (consistent with previous definitions), and $H, W$ are spatial dimensions, we first extract overlapping local patches using an unfold operation:
    $
    U \in \mathbb{R}^{B \times (Ck^2) \times L},
    $
    where $k$ is the kernel size and $L = H_{\text{out}} W_{\text{out}}$ is the number of spatial locations in the output feature map. The dimensions $H_{\text{out}}$ and $W_{\text{out}}$ correspond to the spatial resolution after applying the convolutional parameters (stride, padding, and dilation), following the standard convolution output size formula. Each column $\mathbf{p}_\ell \in \mathbb{R}^{Ck^2}$ corresponds to a vectorized receptive field extracted from all input channels at a given spatial location.
    
    This transformation converts the convolutional operation into a mapping over a set of patch vectors, enabling the application of more expressive nonlinear operators. Instead of applying a linear projection, each patch is processed through a nonlinear mapping:
    $
    \mathbf{z}_\ell = g(\mathbf{p}_\ell),
    $
    where $g(\cdot)$ is implemented as a Kolmogorov--Arnold Network.
    \begin{figure*}[ht]
        \centering
        \begin{adjustbox}{max width=\textwidth}
        \begin{tikzpicture}[
            scale=0.9,
            transform shape,
            node distance=7mm and 8mm,
            every node/.style={font=\scriptsize, align=center},
            block/.style={draw, rounded corners, minimum width=22mm, minimum height=8mm},
            smallblock/.style={draw, rounded corners, minimum width=18mm, minimum height=7mm},
            arrow/.style={->, thick}
        ]
        
        \node[block, fill=blue!10] (patch) {Local Patch\\$\mathbf{p}_\ell \in \mathbb{R}^{Ck^2}$};
        \node[block, fill=green!10, right=of patch] (reshape) {Patch-to-Vector\\representation};
        \node[block, fill=red!10, right=of reshape, minimum width=30mm] (kanlayer) {KAN Layer\\$g(\mathbf{p}_\ell)$};
        
        \node[smallblock, fill=orange!15, above right=4mm and 6mm of kanlayer] (linear) {Factorized\\linear term};
        \node[smallblock, fill=purple!15, below right=4mm and 6mm of kanlayer] (spline) {Spline-based\\nonlinear term};
        
        \node[block, fill=yellow!15, right=28mm of kanlayer] (sum) {Aggregation\\+ LayerNorm};
        \node[block, fill=blue!10, right=of sum] (output) {Output Response\\$\mathbf{z}_\ell \in \mathbb{R}^{D}$};
        
        \draw[arrow] (patch) -- (reshape);
        \draw[arrow] (reshape) -- (kanlayer);
        \draw[arrow] (kanlayer) -- (sum);
        \draw[arrow] (sum) -- (output);
        
        \draw[arrow] (kanlayer.east) -- (linear.west);
        \draw[arrow] (kanlayer.east) -- (spline.west);
        \draw[arrow] (linear.east) -- ([yshift=2mm]sum.west);
        \draw[arrow] (spline.east) -- ([yshift=-2mm]sum.west);
        
        \end{tikzpicture}
        \end{adjustbox}
        \caption{Internal structure of the CKAN transformation applied to each unfolded local patch. Each patch vector is processed by a KAN-based mapping composed of a factorized linear component and a spline-based nonlinear component. Their outputs are aggregated and normalized to produce the transformed patch representation.}
        \label{fig:ckan_internal}
    \end{figure*}
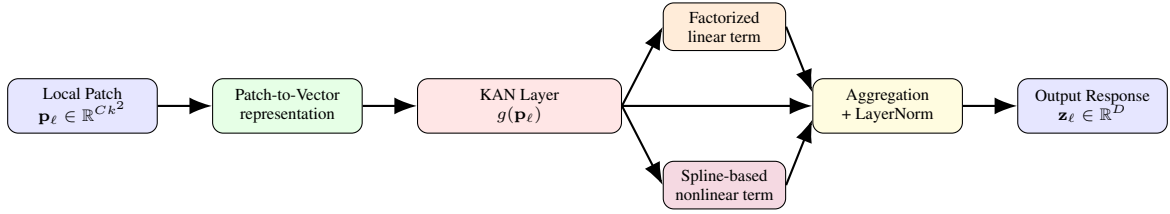

    The KAN module (Figure~\ref{fig:ckan_internal}, \ref{appendix:gen_ckan_operators}) consists of multiple layers, each combining two complementary components. The first is a factorized linear transformation:
    \begin{equation}
    W = \sum_{j=1}^{p} \sum_{k=1}^{s} a_{jk} M_{jk},
    \end{equation}
    where $a_{jk}$ are learnable coefficients and $M_{jk}$ are fixed basis matrices. The second component is a spline-based nonlinear transformation. Each input dimension is processed using a learnable spline expansion:
    \begin{equation}
    \phi(x) = \sum_{m=1}^{D_s} \alpha_m B_m(x),
    \end{equation}
    where $B_m(x)$ are spline basis functions defined over a grid, $\alpha_m$ are learnable coefficients, and $D_s$ denotes the number of spline basis functions (i.e., the spline resolution) used to approximate the nonlinear mapping. The output of each layer is obtained by combining the factorized linear term with the spline-based nonlinear term, followed by normalization:
    \begin{equation}
    y = \mathrm{LayerNorm}(W \sigma(x) + \phi(x)),
    \end{equation}
    where $\sigma(\cdot)$ is a base activation function.
    
    This structure provides a practical realization of the Kolmogorov--Arnold principle, enabling the model to approximate complex multivariate functions through structured nonlinear transformations.
    
    The KAN module is applied independently to each patch vector. In practice, the unfolded tensor is reshaped from $(B, Ck^2, L)$ to $(B  L, Ck^2)$, processed through the KAN network, and then reshaped back to $(B, D, L)$. Finally, the transformed features are rearranged into spatial form: $
    Y \in \mathbb{R}^{D \times H_{\text{out}} \times W_{\text{out}}}.
    $

    \begin{figure*}[ht]
        \centering
        \includegraphics[width=\textwidth]{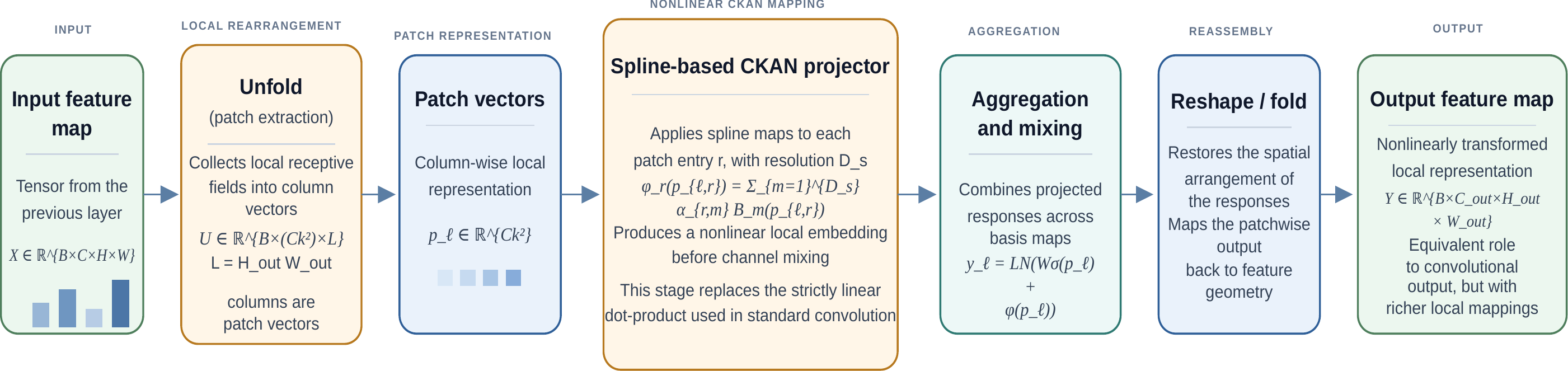}
        \caption{Internal CKAN operator. The input feature map is unfolded into local patches, transformed through spline-based KAN mappings, aggregated, and reshaped back into the output feature map. CKAN substitutes linear kernel–patch multiplication with a spline-driven nonlinear patch projection followed by aggregation and spatial reassembly.}
        \label{fig:ckan_operator}
    \end{figure*}
    
    This unfold--KAN--reshape pipeline defines the CKAN operator (Figure \ref{fig:ckan_operator}), which can be used as a drop-in replacement for standard convolutional layers.

    \subsection{Integration and Practical Considerations}
    
    Direct application of patch-based transformations can be memory-intensive. To address this, we implement a chunked processing strategy that divides the spatial domain into bands, applying the unfold and KAN transformation in smaller segments. This allows the operator to scale to larger images without exceeding memory limits, while preserving exact equivalence to the full operation.In our experimental setup, the chunked implementation was successfully tested on HR images up to $720 \times 1280$ pixels under constrained GPU memory.
    
    The CKAN operator replaces convolutional layers within the residual blocks of the generator, preserving the global SRGAN architecture while modifying the local transformation. Then, by combining patch-space representation with KAN-based nonlinear mappings, the model can capture richer local dependencies and reconstruct high-frequency structures more effectively. This is particularly beneficial in adversarial training, where perceptual realism depends on modeling fine textures.
    
    Overall, the proposed method shifts the focus from increasing network depth to enhancing the expressive power of local operators, introducing a new class of convolutional transformations grounded in approximation theory.
        
	%%%%%%%%%%%%%%%%%%%%%%%%%%%%%%%%%%%%%%%%%%%%%%%%%%%%
    \section{Experiments}
    %%%%%%%%%%%%%%%%%%%%%%%%%%%%%%%%%%%%%%%%%%%%%%%%%%%%
    
    \subsection{Dataset and Training}
    
    We conduct experiments using the DIV2K dataset \cite{div2k}, a widely used benchmark for super-resolution tasks. For computational efficiency, we use a subset consisting of 1600 images in total, divided into 800 images for training and 800 images for validation. All images are resized to a maximum resolution of 720p.
    
    Low-Resolution (LR) inputs are generated from the corresponding High-Resolution (HR) images using bicubic downsampling with a scaling factor of $\times 4$. During training, random HR patches of size $384 \times 384$ are extracted together with their aligned LR counterparts, allowing the model to learn local mappings between degraded and high-resolution regions.
    
    Training is divided into two stages: In the first stage, corresponding to supervised pretraining (SRResNet phase, \ref{appendix:training_objectives}), the generator is optimized using a reconstruction-oriented objective based on content loss (Eq. \ref{appendix:eq:content_loss}), which encourages structural consistency and provides a stable initialization. In the second stage, corresponding to adversarial training (SRGAN phase, \ref{appendix:training_objectives}), the generator is further optimized using a combination of content, pixel, and adversarial losses, while the discriminator is jointly trained to distinguish real HR images from generated SR output.
    
    To improve stability, we employ a controlled early-stopping strategy guided by perceptual quality and constrained by reconstruction fidelity. In particular, model selection during adversarial training considers perceptual improvements while preventing excessive degradation in PSNR.
    
    \subsection{Evaluation Metrics}
    
    We evaluate model performance using Peak Signal-to-Noise Ratio (PSNR), defined as
    \begin{equation}
    \text{PSNR} = 10 \cdot \log_{10} \left( \frac{MAX^2}{\text{MSE}} \right),
    \end{equation}
    where $MAX$ is the maximum possible pixel value and $\text{MSE}$ is the mean squared error between $I^{SR}$ and $I^{HR}$. PSNR is a standard metric for measuring pixel-wise reconstruction fidelity. However, it is well known that PSNR correlates poorly with perceptual quality, often favoring overly smooth reconstructions.
    
    To better assess perceptual quality, we also use the LPIPS metric,
    \begin{equation}
    \text{LPIPS}(x, y) = \sum_{l} w_l \cdot \| \phi_l(x) - \phi_l(y) \|_2^2,
    \end{equation}
    where $\phi_l(\cdot)$ denotes deep feature representations extracted from layer $l$ of a pretrained network and $w_l$ are learned weights. Unlike PSNR, LPIPS better reflects human perceptual judgments and is particularly sensitive to texture and structural realism.
    
    In addition, we report Structural Similarity Index (SSIM) and its multi-scale version (MS-SSIM), which quantify similarity in terms of luminance, contrast, and structure. These metrics are useful for monitoring training behavior and for analyzing the trade-off between reconstruction fidelity and perceptual quality.
    
    In adversarial super-resolution, optimizing for perceptual realism often leads to lower PSNR values. This behavior is expected, since the model prioritizes visually plausible high-frequency details over exact pixel-wise matching. Therefore, a perceptual model should not be judged solely by PSNR, but by a combination of fidelity-oriented and perceptual metrics.
    
    As shown in Fig.~\ref{fig:visual_results}, the adversarial model tends to generate visually sharper textures, whereas the pretrained baseline preserves stronger structural fidelity. This qualitative behavior is consistent with the quantitative trade-off observed in the reported metrics.
    
    \subsection{Training Process}
    
\begin{figure*}[ht]
        \centering
        \includegraphics[width=\textwidth]{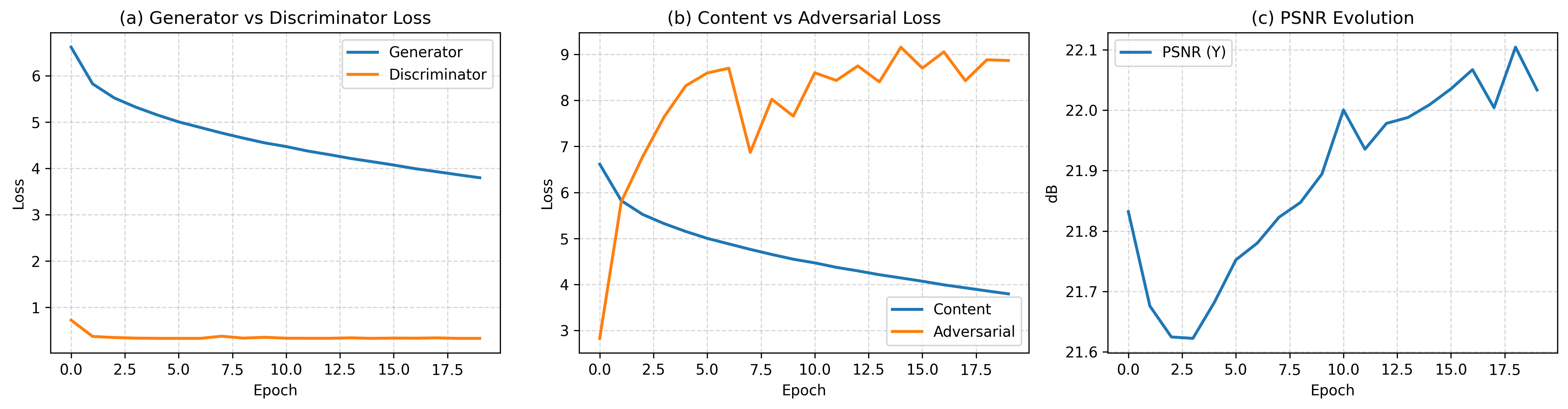}
        \caption{Training dynamics of the SRGAN-CKAN model. (a) Generator and discriminator losses showing stable adversarial training. (b) Evolution of content and adversarial loss components, illustrating the balance between reconstruction and perceptual objectives. (c) PSNR evolution on the luminance channel, where the dashed line indicates the pretrained SRResNet baseline.}
        \label{fig:training_curves}
    \end{figure*}

    Training GANs is inherently unstable due to the adversarial interaction between the generator and the discriminator. Understanding this behavior is essential to evaluate convergence and model performance.     Figure~\ref{fig:training_curves} shows the evolution of losses and metrics during training. The generator loss decreases progressively while the discriminator remains bounded, indicating a stable adversarial process under the selected hyperparameter configuration.
    
    Figure~\ref{fig:training_curves}(b) illustrates the evolution of the content and adversarial loss components. As training progresses, the content loss stabilizes while the adversarial component gradually increases, reflecting the growing influence of perceptual objectives. This behavior indicates a balanced optimization process in which reconstruction fidelity is preserved while encouraging the generation of perceptually realistic textures. At the same time, PSNR remains below the pretrained baseline, which is consistent with the expected fidelity--perception trade-off in adversarial super-resolution.
    
    %%%%%%%%%%%%%%%%%%%%%%%%%%%%%%%%%%%%%%%%%%%%%%%%%%%%
    \subsection{Implementation Details}
    %%%%%%%%%%%%%%%%%%%%%%%%%%%%%%%%%%%%%%%%%%%%%%%%%%%%
    
    Under the paradigm of minimal resources, experiments were conducted on a system equipped with an NVIDIA RTX 4060 GPU (8 GB VRAM), under constrained computational resources. Due to memory limitations, training was performed with a batch size of 1 and HR patches of size $384 \times 384$.

    The model was implemented in PyTorch with CUDA acceleration on a Linux-based system, featuring an AMD Ryzen 9 CPU and 64 GB of RAM. Low-resolution inputs were generated via bicubic downsampling with a scaling factor of $\times 4$. The generator was first pretrained using a reconstruction objective and subsequently fine-tuned in an adversarial setting. Additional implementation details and training configurations are provided in the supplementary material (\ref{appendix:srgan_ckan_architecture}).
    
    %%%%%%%%%%%%%%%%%%%%%%%%%%%%%%%%%%%%%%%%%%%%%%%%%%%%
    \section{Experimental Protocol and Analysis}
    %%%%%%%%%%%%%%%%%%%%%%%%%%%%%%%%%%%%%%%%%%%%%%%%%%%%
    
    A fair comparison with SRGAN requires consistent alignment of the degradation model, evaluation datasets, and metric definitions, as variations in these factors can significantly affect reported results. Therefore, the focus of this study is not absolute numerical superiority, but the controlled characterization of the fidelity--perceptual trade-off induced by adversarial optimization.
    
    The pretrained SRResNet-CKAN model serves as a reconstruction-oriented baseline, while adversarial fine-tuning shifts the model toward improved perceptual quality. This transition reflects the known behavior of GAN-based super-resolution, where perceptual realism is typically achieved at the cost of slight degradation in distortion-based metrics such as PSNR and MS-SSIM.
    
    To ensure reproducibility, we adopt a structured training protocol consisting of supervised pretraining, controlled adversarial fine-tuning, and monitored stopping criteria. This setup allows systematic analysis of training dynamics rather than relying on purely heuristic optimization.
    
    From an architectural perspective, the proposed formulation remains extensible. The \texttt{unfold}+$\mathcal{P}$ representation explicitly models local patches as feature vectors, providing a foundation for future nonlinear spline-based mappings inspired by Kolmogorov--Arnold theory.
    
    Overall, the proposed framework establishes a controlled experimental setting for analyzing adversarial super-resolution with CKAN-based generators, emphasizing interpretability and reproducibility.
    
  %%%%%%%%%%%%%%%%%%%%%%%%%%%%%%%%%%%%%%%%%%%%%%%%%%%%
    \subsection{Quantitative Results}
    %%%%%%%%%%%%%%%%%%%%%%%%%%%%%%%%%%%%%%%%%%%%%%%%%%%%
    
    Table~\ref{tab:quantitative_results} summarizes the quantitative evaluation of the proposed SRGAN--CKAN model. The comparison includes both the supervised pretraining stage (SRResNet--CKAN) and the adversarial fine-tuning stage, as well as a conventional SRGAN baseline.
    
    From SRResNet--CKAN to SRGAN--CKAN, the PSNR(Y) decreases by only 0.055 dB and MS-SSIM(Y) by 0.0022, indicating that structural fidelity is largely preserved during adversarial training. At the same time, LPIPS improves from 0.2124 to 0.1947, reflecting enhanced perceptual quality.
    
    In comparison with the conventional SRGAN baseline, the proposed SRGAN--CKAN achieves significant improvements across all metrics, with higher PSNR and MS-SSIM, and a substantial reduction in LPIPS. This suggests a more favorable trade-off between reconstruction fidelity and perceptual realism.
    
    These results indicate that the CKAN-based generator modifies adversarial training dynamics, enabling perceptual refinement without the severe distortion penalties typically observed in conventional SRGAN, where adversarial optimization often leads to a significant drop in PSNR.
    
    The improvements are consistent with the qualitative results shown in Figures~\ref{fig:sr_comparison},~\ref{fig:visual_comparison},~\ref{fig:visual_results} and~\ref{fig:conv_vs_ckan}.

    \begin{table}[ht]
        \centering
        \caption{Quantitative comparison of SRResNet--CKAN, SRGAN (convolutional), and SRGAN--CKAN. PSNR and MS-SSIM are computed on the luminance channel. Lower LPIPS indicates better perceptual quality.}
        \label{tab:quantitative_results}
        \begin{tabular}{lccc}
        \toprule
        \textbf{Model} & \textbf{PSNR-Y (dB)} & \textbf{MS-SSIM-Y} & \textbf{LPIPS} \\
        \midrule
        SRResNet--CKAN & 24.9572 & 0.9445 & 0.2124 \\
        SRGAN (conv)   & 20.07   & 0.8550 & 0.3226 \\
        \textbf{SRGAN--CKAN} & \textbf{24.9021} & \textbf{0.9422} & \textbf{0.1947} \\
        \bottomrule
        \end{tabular}
    \end{table}
    \vspace{-5pt}
    %%%%%%%%%%%%%%%%%%%%%%%%%%%%%%%%%%%%%%%%%%%%%%%%%%%%
    %\section{Discussion}
    %%%%%%%%%%%%%%%%%%%%%%%%%%%%%%%%%%%%%%%%%%%%%%%%%%%%
    
    A central observation of this work is the inherent trade-off between reconstruction fidelity and perceptual quality in single-image super-resolution. While pixel-wise optimization yields high PSNR values, it often produces overly smooth results, whereas adversarial training enhances high-frequency details at the cost of distortion metrics.
    
    The proposed SRGAN-CKAN model achieves a favorable balance between these objectives. The minimal degradation in PSNR and MS-SSIM, combined with improved LPIPS, indicates that perceptual quality is enhanced while structural information is preserved. In addition, adversarial fine-tuning exhibits stable behavior, as the transition from pretraining does not lead to abrupt degradation in reconstruction metrics.
    
    The introduction of CKAN operators modifies local feature transformations by replacing standard convolutions with patch-based mappings, enabling more expressive modeling of nonlinear dependencies, particularly in textured regions. This increased expressivity is reflected in sharper textures and improved perceptual similarity, while also contributing to more stable adversarial optimization.

    Finally we provide a contextual reference to state-of-the-art methods such as RCAN \citep{zhang2018imagesuperresolutionusingdeep} and SR3 \citep{saharia2021imagesuperresolutioniterativerefinement}, which rely on deep architectures and large-scale training (often exceeding $10^5$ iterations) on high-end GPU hardware. In contrast, SRGAN--CKAN is trained under constrained conditions using a low-range GPU, a batch size of 1, and a limited number of epochs, yet achieves competitive perceptual quality with improved MS-SSIM and LPIPS while maintaining stable reconstruction performance.some state-of-the-art methods report higher PSNR due to stronger optimization toward distortion-based objectives, often at the expense of perceptual quality. This behavior reflects the perception-distortion trade-off in super-resolution and should be interpreted in the context of the target objective rather than as a direct performance gap.
    
    %Future work will focus on improving efficiency and scalability, including factorized or low-rank variants, and evaluating the proposed method on standardized benchmarks such as Set5, Set14, and Urban100. Exploring multi-scale architectures and integrating perceptual metrics directly into the training objective also represent promising directions.

    %%%%%%%%%%%%%%%%%%%%%%%%%%%%%%%%%%%%%%%%%%%%%%%%%%%%
    %\section{Limitations}
    %%%%%%%%%%%%%%%%%%%%%%%%%%%%%%%%%%%%%%%%%%%%%%%%%%%%
    
    %This work is subject to several limitations. Experiments were conducted under constrained computational resources, limiting batch size and image resolution. Evaluation was performed on a restricted dataset, which may affect generalization. Additionally, the method was not compared against highly optimized state-of-the-art models under standardized benchmarks, as the focus was on analyzing the behavior of the proposed operator.
    
	%%%%%%%%%%%%%%%%%%%%%%%%%%%%%%%%%%%%%%%%%%%%%%%%%%%%
    \section{Conclusion}
    %%%%%%%%%%%%%%%%%%%%%%%%%%%%%%%%%%%%%%%%%%%%%%%%%%%%
    
    In this work, we introduced SRGAN--CKAN, a hybrid super-resolution framework that integrates Kolmogorov--Arnold-inspired operators into an adversarial learning setting.

    By reformulating convolution as a nonlinear patch-based transformation, the proposed CKAN operator increases the expressive capacity of local feature processing, enabling improved modeling of high-frequency textures and perceptual details. In particular, the integration of spline-based nonlinear mappings allows the model to capture complex local relationships beyond linear combinations.
    
    Experimental results demonstrate that the proposed approach enhances perceptual quality while maintaining reconstruction fidelity, achieving a favorable balance between distortion-based and perceptual metrics under adversarial training.
    
    Beyond empirical results, this work highlights the potential of integrating functional representations into deep learning architectures. In particular, CKAN-based operators provide a promising direction for designing more expressive and flexible models for image restoration.
    
    Future work will focus on improving the computational efficiency and scalability, extending evaluation to standardized benchmarks and exploring the integration of lightweight attention mechanisms to further enhance feature selection and adaptive processing.
    \newpage
	% ============================
	%   BIBLIOGRAFÍA
	% ============================

    \newpage
    \appendix
    \section{Supplementary Material}
    \subsection{Kolmogorov--Arnold Representation Theorem}
    
    To address the limitations of linear local operators, we consider the Kolmogorov--Arnold representation theorem, a fundamental result in approximation theory. This theorem states that any continuous multivariate function $f: \mathbb{R}^n \rightarrow \mathbb{R}$ can be represented as:
    \begin{equation*}
    f(x_1, \dots, x_n) = \sum_{q=1}^{2n+1} \Phi_q \left( \sum_{p=1}^{n} \phi_{p,q}(x_p) \right),
    \end{equation*}
    where $\phi_{p,q}(\cdot)$ are univariate nonlinear functions and $\Phi_q(\cdot)$ are outer aggregation functions.
    
    This result is particularly relevant in the context of deep learning, as it provides a constructive way to decompose complex multivariate interactions into compositions of simpler one-dimensional transformations. In contrast to traditional neural networks, which rely on linear projections followed by nonlinear activations, the Kolmogorov--Arnold formulation emphasizes nonlinear transformations at the level of individual input dimensions.
    
    From a representation standpoint, this suggests that highly nonlinear mappings—such as those required in super-resolution—can be efficiently approximated by combining univariate nonlinear functions with learned aggregation mechanisms. This perspective motivates the design of Kolmogorov--Arnold Networks (KANs), where learnable functions replace fixed linear weights.
    
    In the context of image super-resolution, each local patch can be viewed as a high-dimensional input vector, and the reconstruction process can be interpreted as learning a complex multivariate function over this space. The Kolmogorov--Arnold theorem therefore provides a theoretical foundation for replacing linear convolutional kernels with nonlinear operators that act directly on local patches.
    
    This insight directly motivates the CKAN operator introduced in this work, where local receptive fields are processed through structured nonlinear mappings inspired by Kolmogorov--Arnold decompositions, enabling richer and more expressive modeling of local image structures.
        
    %%%%%%%%%%%%%%%%%%%%%%%%%%%%%%%%%%%%%%%%%%%%%%%%%%%%
    \subsection{SRGAN--CKAN Architecture}
    \label{appendix:srgan_ckan_architecture}
    %%%%%%%%%%%%%%%%%%%%%%%%%%%%%%%%%%%%%%%%%%%%%%%%%%%%
        
        From a statistical viewpoint, super-resolution requires both global and local modeling capabilities. Global priors are enforced through adversarial training, which encourages outputs to match the distribution of natural images, while local flexibility is achieved through expressive operators capable of modeling nonlinear relationships between neighboring pixels.
        
        Let $I^{HR} \in \mathbb{R}^{3 \times H \times W}$ be a high-resolution image and assume a degradation process:
        \begin{equation}
        I^{LR} = \mathcal{D}(I^{HR}) = (I^{HR} \ast k)\downarrow_s + n.
        \end{equation}
        The objective is to learn a generator $G_\theta$ such that:
        \begin{equation}
        \hat{I}^{HR} = G_\theta(I^{LR}), \quad \hat{I}^{HR} \approx I^{HR}.
        \end{equation}
        
        Given a batch $I^{LR} \in \mathbb{R}^{B \times 3 \times h \times w}$, the generator produces:
        \[
        \hat{I}^{HR} \in \mathbb{R}^{B \times 3 \times (s h) \times (s w)},
        \]
        where $s$ is the upscaling factor. During training, both real and generated images are processed by a discriminator $D_\phi$, which assigns higher scores to real samples than to generated ones. This adversarial mechanism introduces a perceptual prior, allowing the model to favor visually plausible reconstructions over strictly pixel-wise accurate ones.
        
        \subsubsection{Generator with CKAN operators}
        \label{appendix:gen_ckan_operators}

        The generator follows a SRResNet-type architecture composed of an initial projection layer, a sequence of residual blocks, a global skip connection, and PixelShuffle-based upsampling. Each residual block computes:
        \begin{equation}
        \mathbf{y} = \mathbf{x} + \mathcal{F}(\mathbf{x}),
        \end{equation}
        where $\mathcal{F}$ represents a local transformation.
        
        In standard convolutional networks, $\mathcal{F}$ is implemented as a linear operator applied to local patches. In contrast, the proposed model replaces this operation with a nonlinear patch-based transformation defined through the CKAN operator.
        
        Given an input feature map $X \in \mathbb{R}^{B \times C \times H \times W}$, local patches are first extracted using an unfold operation:
        \begin{equation}
        U = \texttt{unfold}(X) \in \mathbb{R}^{B \times (C k_H k_W) \times L},
        \end{equation}
        where each column $\mathbf{p}_\ell \in \mathbb{R}^{C k_H k_W}$ represents a vectorized receptive field and $L$ is the number of spatial locations.
        
        Each patch is then processed independently through a nonlinear mapping:
        \begin{equation}
        \mathbf{z}_\ell = \mathcal{P}(\mathbf{p}_\ell),
        \end{equation}
        where $\mathcal{P}$ is defined using a Kolmogorov--Arnold Network (KAN).
        
        Following the Kolmogorov--Arnold representation theorem, a multivariate function can be expressed as a sum of univariate transformations. In the context of CKAN, this leads to a functional mapping of the form:
        \begin{equation}
        \mathcal{P}(\mathbf{p}_\ell) = \sum_{q=1}^{Q} \Phi_q \left( \sum_{i=1}^{d} \phi_{i,q}(p_{\ell,i}) \right),
        \end{equation}
        where $d = C k_H k_W$, $\phi_{i,q}$ are learnable univariate functions, and $\Phi_q$ are aggregation functions.
        
        In practice, these functions are implemented using spline-based parameterizations over a learnable grid, following the LTBs-KAN formulation \cite{merinmartinez2026ltbskanlineartimebsplineskolmogorovarnold}. This allows efficient approximation of nonlinear mappings with linear computational complexity in the number of input dimensions.
        
        The final operator can therefore be interpreted as replacing the linear projection:
        \begin{equation}
        y_{i,j} = \mathbf{w}^\top \mathbf{p}_\ell
        \end{equation}
        with a nonlinear functional mapping:
        \begin{equation}
        y_{i,j} = \mathcal{P}(\mathbf{p}_\ell),
        \end{equation}
        which enables richer modeling of local interactions and improves the reconstruction of high-frequency details.
        
        The transformed patches are then rearranged into spatial form, producing the output feature map. To control memory usage, the unfold operation can be applied in bands without affecting the resulting output.
        
        \subsubsection{Training objectives}
        \label{appendix:training_objectives}

    Training is performed in two stages: a reconstruction-oriented pretraining phase and an adversarial fine-tuning phase.

    \begin{figure*}[t]
        \centering
        \includegraphics[width=\textwidth]{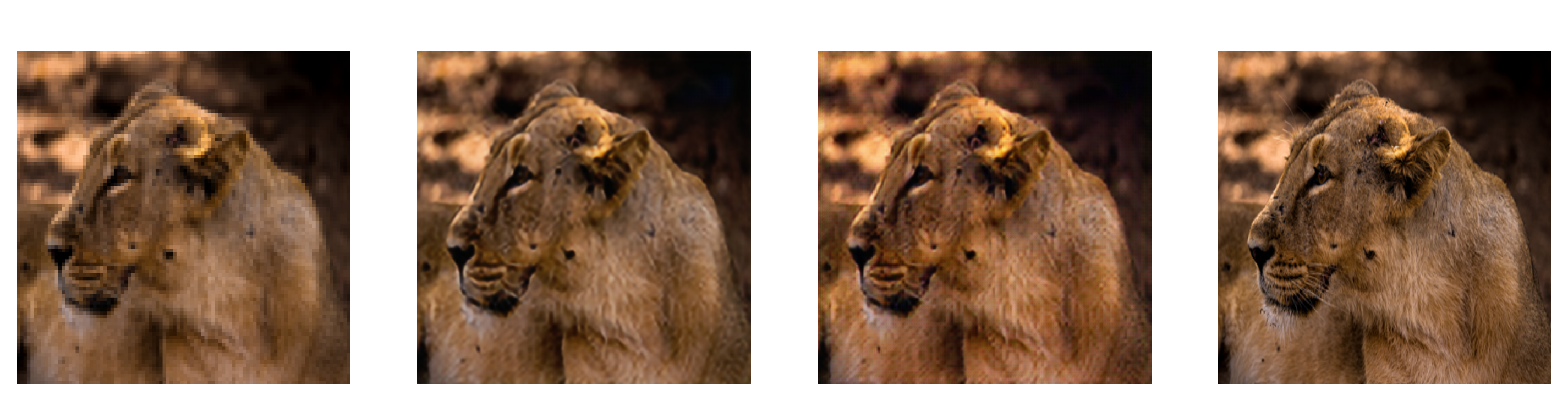}
        \vspace{2mm}
        \includegraphics[
        height=0.2\textheight,
        keepaspectratio
        ]{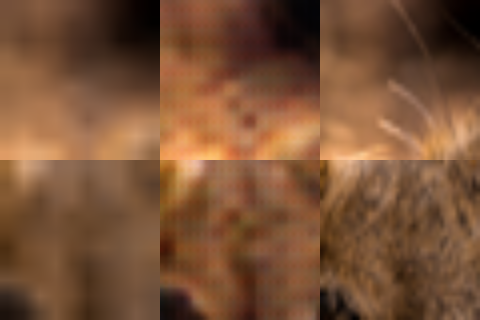}
        \caption{Qualitative comparison of super-resolution results. From left to right: low-resolution input (LR), reconstruction using SRResNet-CKAN, reconstruction using SRGAN-CKAN, and ground-truth high-resolution image (HR). The adversarial model produces sharper textures and enhanced perceptual detail, while the reconstruction-based model yields smoother outputs.}
        \label{fig:visual_comparison}
    \end{figure*}
    
    \paragraph{Generator loss}
    
    During adversarial training, the generator is optimized using a weighted combination of perceptual, adversarial, and pixel-wise losses:
    \begin{equation}
    \label{appendix:eq:content_loss}
    \mathcal{L}_G = \lambda_{adv}\mathcal{L}_{adv} + \lambda_{perc}\mathcal{L}_{perc} + \lambda_{pix}\mathcal{L}_{pix}.
    \end{equation}
    
    The adversarial loss encourages the generator to produce realistic images:
    \begin{equation}
    \mathcal{L}_{adv} = \text{BCEWithLogits}(D(\hat{I}^{HR}), 1).
    \end{equation}
    
    The perceptual loss is defined in feature space using a pretrained network (e.g., VGG):
    \begin{equation}
    \mathcal{L}_{perc} = \left\| \phi(I^{HR}) - \phi(\hat{I}^{HR}) \right\|_2^2,
    \end{equation}
    where $\phi(\cdot)$ denotes feature extraction from a fixed deep network.
    
    The pixel-wise reconstruction loss ensures fidelity to the ground truth:
    \begin{equation}
    \mathcal{L}_{pix} = \left\| I^{HR} - \hat{I}^{HR} \right\|_1.
    \end{equation}
    
    \paragraph{Discriminator loss}
    
    The discriminator is trained to distinguish real from generated samples:
    \begin{equation}
        \begin{aligned}
        \mathcal{L}_D
        &= \text{BCEWithLogits}(D(I^{HR}), 1) \\
        &\quad + \text{BCEWithLogits}(D(\hat{I}^{HR}), 0).
        \end{aligned}
    \end{equation}

    \begin{figure*}[ht]
            \centering
            \includegraphics[
                width=\textwidth,
                height=0.3\textheight,
                keepaspectratio
            ]{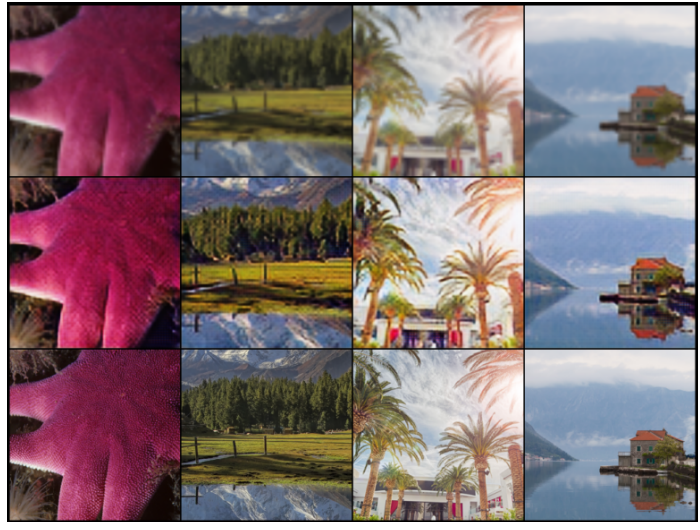}
            \caption{Qualitative comparison among LR input, SRResNet-CKAN output, SRGAN-CKAN output, and HR target (Columns in the matrix of images). The figure illustrates the trade-off between reconstruction fidelity and perceptual sharpness.}
            \label{fig:visual_results}
    \end{figure*}

    \paragraph{Interpretation}
    
    This formulation enforces a balance between reconstruction fidelity and perceptual realism. The pixel and perceptual losses preserve structural information, while the adversarial term encourages the generation of high-frequency details. In the context of CKAN-based operators, this combination allows the model to exploit its increased local expressivity to produce sharper and more realistic textures without significant degradation in distortion-based metrics.

    %%%%%%%%%%%%%%%%%%%%%%%%%%%%%%%%%%%%%%%%%%%%%%%%%%%%
    \section{Computational Complexity LBTs-KAN}
    \label{appendix:ComputationaComplexityLBTs-KAN}
    %%%%%%%%%%%%%%%%%%%%%%%%%%%%%%%%%%%%%%%%%%%%%%%%%%%%
        
        Based on the algorithms at Merin-Martinez et al. \cite{merinmartinez2026ltbskanlineartimebsplineskolmogorovarnold} the computational cost of the proposed CKAN operator differs from standard convolution due to its explicit patch-based formulation and nonlinear projection mechanism. A conventional convolution layer has complexity:
        \begin{equation}
        \mathcal{O}(B \ C_{\text{out}}  C  k^2  H_{\text{out}} W_{\text{out}}),
        \end{equation}
        where $B$ is the batch size, $C$ and $C_{\text{out}}$ are input and output channels, and $k$ is the kernel size.
        
     Let the input tensor be
    \begin{equation*}
    x \in \mathbb{R}^{B \times C_{\mathrm{in}} \times H \times W}.
    \end{equation*}
    
    Thus, we have the next equations $k = k_H k_W, K = C_{\mathrm{in}} k,$ and $ L = H_{\mathrm{out}} W_{\mathrm{out}}$, where:
    \begin{align*}
    H_{\mathrm{out}} &= \left\lfloor \frac{H + 2p_H - d_H(k_H - 1) - 1}{s_H} + 1 \right\rfloor, \\
    W_{\mathrm{out}} &= \left\lfloor \frac{W + 2p_W - d_W(k_W - 1) - 1}{s_W} + 1 \right\rfloor. 
   \end{align*}
    
    Thus, each extracted patch has dimension $K = C_{\mathrm{in}} k_H k_W$. Now, we investigate the unfold (Im2Col) Complexity, the operation \texttt{F.unfold} produces:
    \begin{equation*}
    \texttt{col} \in \mathbb{R}^{B \times K \times L}.
    \end{equation*}
    
    Therefore, the the computational cost of this is:
    \begin{equation*}
    \mathcal{T}_{\text{unfold}} =
    O(B K L)=
    O\left(
    B C_{\mathrm{in}} k_H k_W H_{\mathrm{out}} W_{\mathrm{out}}
    \right)
    \end{equation*}
    Moreover, the projection module maps:
    \begin{equation}
    \mathbb{R}^{B \times K \times L}
    \longrightarrow
    \mathbb{R}^{B \times C_{\mathrm{out}} \times L}.
    \end{equation}
    
    Now, we denote its cost as $\mathcal{T}_{\text{proj}}(B,K,L,C_{\mathrm{out}})$. Thus, the total complexity of the layer is:
    \begin{equation}
    \mathcal{T}_{\text{KANConv}} =
    O(B K L) + \mathcal{T}_{\text{proj}}(B,K,L,C_{\mathrm{out}})
    \end{equation}
    
    If the projection operates independently per patch, then:
    \begin{equation}
    \mathcal{T}_{\text{proj}} =
    O\left(B L  \mathcal{C}_{\text{patch}}\right),
    \end{equation}
    
    where $\mathcal{C}_{\text{patch}}$ is the cost of processing a single patch.
    
    Hence:
    \begin{equation*}
    \mathcal{T}_{\text{KANConv}} =
    O\left(B L (K + \mathcal{C}_{\text{patch}})\right)
    \end{equation*}
    
    Now, if the projector is linear:
    \begin{equation*}
    \mathcal{C}_{\text{patch}} = O(K C_{\mathrm{out}}),
    \end{equation*}
    
    then:
    \begin{equation*}
    \mathcal{T}_{\text{KANConv}} =
    O(B L K C_{\mathrm{out}})
    \end{equation*}
    
    which matches the complexity of standard convolution via \text{im2col}. If the projector is a KAN block with spline resolution $D$, then:
    \begin{equation}
    \mathcal{C}_{\text{patch}} =
    O\left(
    \sum_{\ell=1}^{m} d_{\ell-1} d_{\ell} D
    \right),
    \end{equation}
    
    and:
    \begin{equation}
    \mathcal{T}_{\text{KANConv}} =
    O\left(
    B L \sum_{\ell=1}^{m} d_{\ell-1} d_{\ell} D
    \right)
    \end{equation}

    Now, for chunked computation,  $L > \texttt{chunk\_pixels}$, the computation is split into $q$ chunks:
    \begin{equation*}
    \sum_{i=1}^{q} L_i = L.
    \end{equation*}
    
    Each chunk has cost $O(B K L_i) + \mathcal{T}_{\text{proj}}(B,K,L_i,C_{\mathrm{out}})$ and,  summing over all chunks:
    \begin{equation*}
    \mathcal{T}_{\text{chunked}} =
    O(B K L) + \mathcal{T}_{\text{proj}}(B,K,L,C_{\mathrm{out}})
    \end{equation*}
    
    Thus, chunking does not change the asymptotic time complexity. Finally in Full Computation
    \begin{equation}
    \mathcal{S}_{\text{full}} = O(B K L)
    \end{equation}
 Finally, 
    \begin{equation}
    \mathcal{S}_{\text{chunked}} =
    O(B K  \texttt{chunk\_pixels})
    \end{equation}

    The proposed chunked KAN convolution preserves the asymptotic computational complexity of the standard im2col-based formulation while significantly reducing peak memory usage. This enables scalable training on high-resolution inputs without altering the theoretical time complexity of the model.

    \subsection{SRGAN (conv) Analysis}
    \label{sec:supplementary_srgan_conv}
    
    To support the quantitative comparison presented in Section~\ref{tab:quantitative_results}, we provide additional qualitative and training dynamics analysis for the conventional SRGAN (conv) baseline.
    
    \subsubsection{Qualitative Analysis}

\begin{figure*}[ht]
        \centering
       \includegraphics[
                width=\textwidth,
                height=0.35\textheight,
                keepaspectratio
            ]{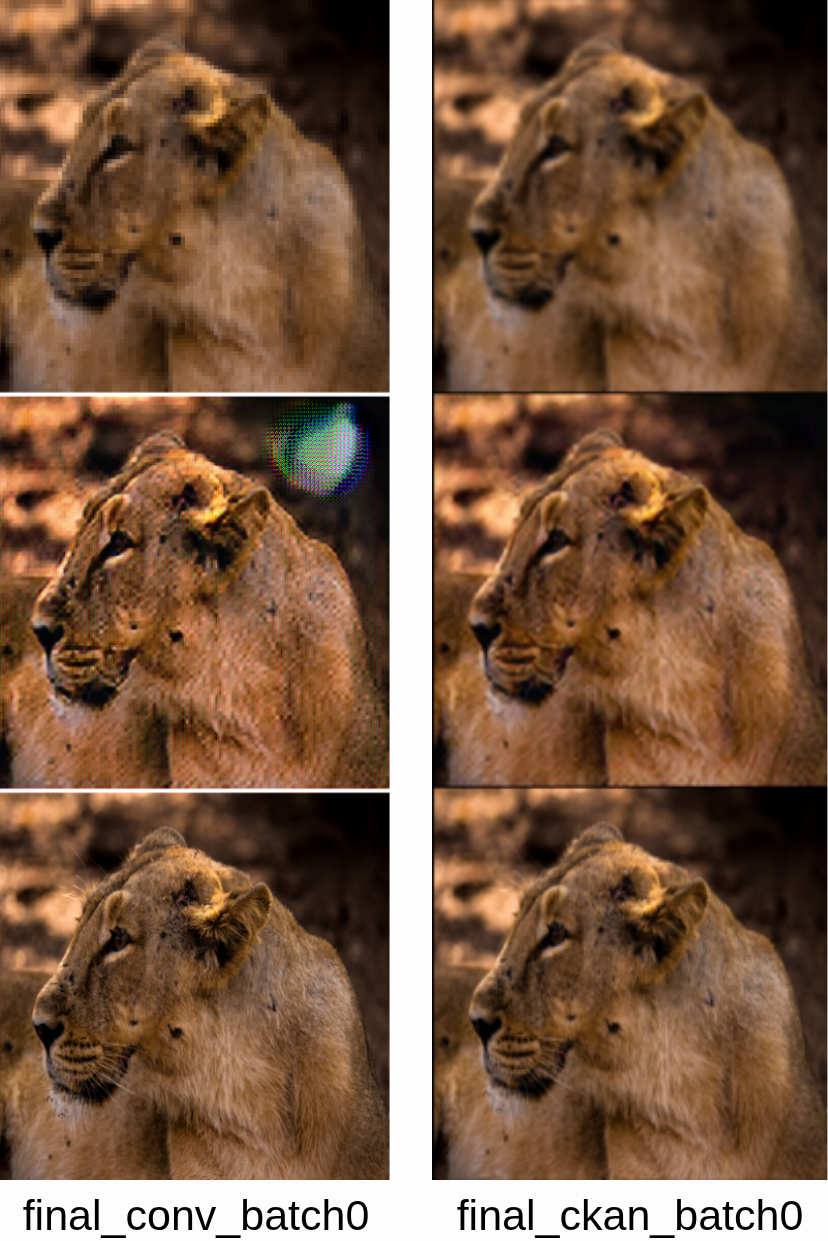}
        \caption{Qualitative comparison between the convolutional SRGAN baseline and the proposed SRGAN-CKAN model. The left column corresponds to the convolutional generator, while the right column shows the CKAN-based generator. In the highlighted examples, the convolutional baseline exhibits visible artifacts and less stable texture reconstruction, whereas the proposed model produces cleaner and more consistent outputs.}
        \label{fig:conv_vs_ckan}
    \end{figure*}

    Figure~\ref{fig:conv_vs_ckan} provides a direct qualitative comparison between the convolutional SRGAN baseline and the proposed SRGAN-CKAN model. Although both approaches recover the global structure of the scene, the convolutional baseline introduces visible artifacts in challenging regions, while the CKAN-based model produces visually cleaner and more stable reconstructions. This suggests that the proposed formulation improves the robustness of local feature transformation during adversarial super-resolution.While the model with the conventional SRGAN enhances high-frequency textures compared to SRResNet, it also introduces visible artifacts and noise patterns in certain regions. These artifacts are particularly noticeable in homogeneous or low-texture areas, where adversarial optimization tends to amplify spurious high-frequency signals.
    
    This behavior is consistent with the observed degradation in distortion-based metrics (PSNR and MS-SSIM) and supports the interpretation that conventional SRGAN prioritizes perceptual sharpness at the expense of structural fidelity.
    
    \subsubsection{Training Dynamics}
    
\begin{figure}[ht]
        \centering
        \includegraphics[width=\linewidth]{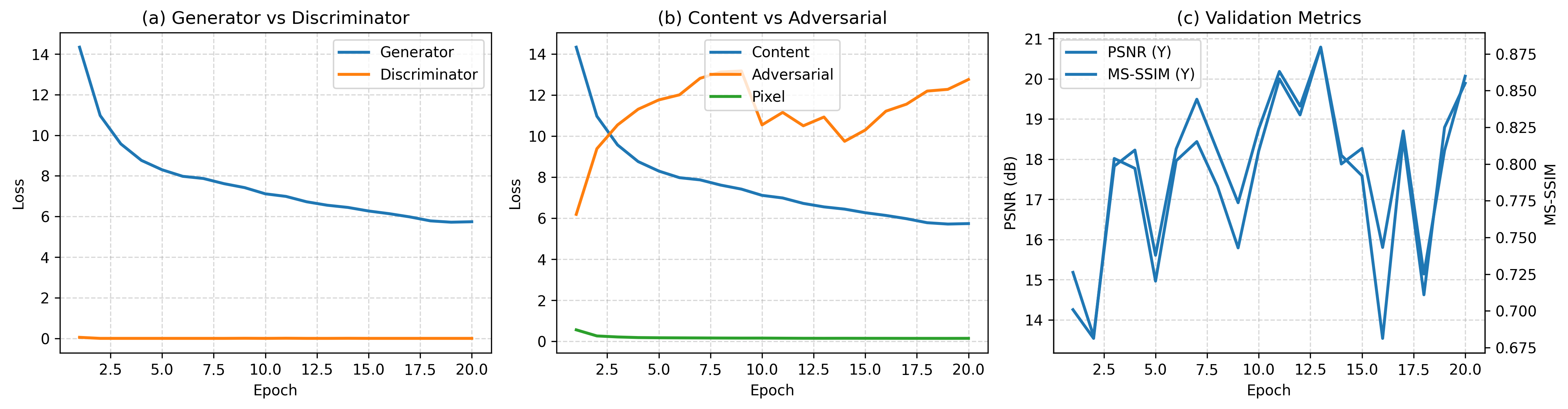}
        \caption{
        Training curves for SRGAN (conv), including generator vs discriminator losses, loss decomposition, and validation metrics across epochs.
        }
        \label{fig:supp_srgan_training}
    \end{figure}

    Figure~\ref{fig:supp_srgan_training} shows the training dynamics of the conventional SRGAN model, including generator and discriminator losses, as well as the evolution of content, adversarial, and pixel losses.
    
    The curves reveal the characteristic behavior of adversarial training: while the generator loss decreases gradually, the adversarial component remains relatively high, indicating strong pressure from the discriminator. This imbalance contributes to the observed trade-off between perceptual quality and reconstruction fidelity.
    
    Additionally, the validation metrics exhibit noticeable fluctuations across epochs, reflecting the inherent instability of GAN optimization. These dynamics further explain the degradation in PSNR and MS-SSIM reported in Table~\ref{tab:quantitative_results}.


\begin{thebibliography}{16}
\providecommand{\natexlab}[1]{#1}
\providecommand{\url}[1]{\texttt{#1}}
\expandafter\ifx\csname urlstyle\endcsname\relax
  \providecommand{\doi}[1]{doi: #1}\else
  \providecommand{\doi}{doi: \begingroup \urlstyle{rm}\Url}\fi

\bibitem[Dong et~al.(2014)Dong, Loy, He, and Tang]{dong2014srcnn}
Chao Dong, Chen~Change Loy, Kaiming He, and Xiaoou Tang.
\newblock Learning a deep convolutional network for image super-resolution.
\newblock In \emph{ECCV}, 2014.

\bibitem[Kim et~al.(2016)Kim, Lee, and Lee]{kim2016vdsr}
Jiwon Kim, Jung~Kwon Lee, and Kyoung~Mu Lee.
\newblock Accurate image super-resolution using very deep convolutional networks.
\newblock In \emph{CVPR}, 2016.

\bibitem[Wang and et~al.(2021)]{srSurveyModern}
Xintao Wang and et~al.
\newblock Deep learning for image super-resolution: A survey.
\newblock \emph{IEEE TPAMI}, 2021.

\bibitem[Gonzalez and Woods(2018)]{gonzalezwoods}
Rafael~C. Gonzalez and Richard~E. Woods.
\newblock \emph{Digital Image Processing}.
\newblock Pearson, 2018.

\bibitem[Jain(1989)]{jain1989fundamentals}
Anil~K. Jain.
\newblock \emph{Fundamentals of Digital Image Processing}.
\newblock Prentice Hall, 1989.

\bibitem[Ledig et~al.(2017)Ledig, Theis, Husz{\'a}r, Caballero, Cunningham, Acosta, Aitken, Tejani, Totz, Wang, and Shi]{ledig2017}
Christian Ledig, Lucas Theis, Ferenc Husz{\'a}r, Jose Caballero, Andrew Cunningham, Alejandro Acosta, Andrew Aitken, Alykhan Tejani, Johannes Totz, Zehan Wang, and Wenzhe Shi.
\newblock Photo-realistic single image super-resolution using a generative adversarial network.
\newblock In \emph{CVPR}, 2017.

\bibitem[Wang et~al.(2018)Wang, Yu, Wu, Gu, Liu, Dong, Qiao, and Loy]{esrgan2018}
Xintao Wang, Ke~Yu, Shixiang Wu, Jinjin Gu, Yihao Liu, Chao Dong, Yu~Qiao, and Chen~Change Loy.
\newblock Esrgan: Enhanced super-resolution generative adversarial networks.
\newblock In \emph{ECCV Workshops}, 2018.

\bibitem[Goodfellow et~al.(2014)Goodfellow, Pouget-Abadie, Mirza, Xu, Warde-Farley, Ozair, Courville, and Bengio]{goodfellow2014gan}
Ian Goodfellow, Jean Pouget-Abadie, Mehdi Mirza, Bing Xu, David Warde-Farley, Sherjil Ozair, Aaron Courville, and Yoshua Bengio.
\newblock Generative adversarial nets.
\newblock In \emph{NeurIPS}, 2014.

\bibitem[Liu et~al.(2024)Liu, Wang, Vaidya, Ruehle, Halverson, Solja{\v{c}}i{\'c}, and Tegmark]{kan2024}
Ziming Liu, Yijun Wang, Varun Vaidya, Fabian Ruehle, James Halverson, Marin Solja{\v{c}}i{\'c}, and Max Tegmark.
\newblock Kan: Kolmogorov--arnold networks.
\newblock \emph{arXiv preprint arXiv:2406.13155}, 2024.

\bibitem[De~Boor et~al.(1993)De~Boor, H{\"o}llig, and Riemenschneider]{deboor1993box}
Carl De~Boor, Klaus H{\"o}llig, and Sherman Riemenschneider.
\newblock \emph{Box Splines}.
\newblock Springer, 1993.

\bibitem[Lim et~al.(2017)Lim, Son, Kim, Nah, and Lee]{lim2017enhanced}
Bee Lim, Sanghyun Son, Heewon Kim, Seungjun Nah, and Kyoung~Mu Lee.
\newblock Enhanced deep residual networks for single image super-resolution.
\newblock In \emph{CVPR Workshops}, 2017.

\bibitem[Zhang et~al.(2018)Zhang, Li, Li, Wang, Zhong, and Fu]{zhang2018imagesuperresolutionusingdeep}
Yulun Zhang, Kunpeng Li, Kai Li, Lichen Wang, Bineng Zhong, and Yun Fu.
\newblock Image super-resolution using very deep residual channel attention networks, 2018.
\newblock URL \url{https://arxiv.org/abs/1807.02758}.

\bibitem[Liang et~al.(2021)Liang, Cao, Sun, Zhang, Van~Gool, and Timofte]{liang2021swinir}
Jingyun Liang, Jiezhang Cao, Guolei Sun, Kai Zhang, Luc Van~Gool, and Radu Timofte.
\newblock Swinir: Image restoration using swin transformer.
\newblock In \emph{ICCV Workshops}, 2021.

\bibitem[Saharia et~al.(2021)Saharia, Ho, Chan, Salimans, Fleet, and Norouzi]{saharia2021imagesuperresolutioniterativerefinement}
Chitwan Saharia, Jonathan Ho, William Chan, Tim Salimans, David~J. Fleet, and Mohammad Norouzi.
\newblock Image super-resolution via iterative refinement, 2021.
\newblock URL \url{https://arxiv.org/abs/2104.07636}.

\bibitem[Merin-Martinez et~al.(2026)Merin-Martinez, Mendez-Vazquez, and Rodriguez-Tello]{merinmartinez2026ltbskanlineartimebsplineskolmogorovarnold}
Eduardo~Said Merin-Martinez, Andres Mendez-Vazquez, and Eduardo Rodriguez-Tello.
\newblock Ltbs-kan: Linear-time b-splines kolmogorov-arnold networks, 2026.
\newblock URL \url{https://arxiv.org/abs/2604.22034}.

\bibitem[Agustsson and Timofte(2017)]{div2k}
Eirikur Agustsson and Radu Timofte.
\newblock Ntire 2017 challenge on single image super-resolution: Dataset and study.
\newblock In \emph{CVPR Workshops}, 2017.

\end{thebibliography}
\end{document}